\documentclass[preprint,12pt,authoryear]{elsarticle}

\usepackage[english]{babel}
\usepackage[utf8x]{inputenc}
\usepackage[T1]{fontenc}

\usepackage[a4paper,top=3cm,bottom=2cm,left=3cm,right=3cm,marginparwidth=1.75cm]{geometry}

\usepackage{amsmath}
\usepackage{graphicx}
\usepackage{caption}
\usepackage{subcaption}
\usepackage[colorinlistoftodos]{todonotes}
\usepackage[colorlinks=true, allcolors=blue]{hyperref}
\usepackage{lineno}
\usepackage{pgfplots}
\usepackage{filecontents}

\journal{Cytometry Part A}

\begin{document}

\begin{frontmatter}

\title{Semantic segmentation of mFISH images using convolutional networks}

\author[label1]{Esteban Pardo}
\author[label2]{José Mário T Morgado}
\author[label1]{Norberto Malpica}

\address[label1]{Medical Image Analysis and Biometry Lab, Universidad Rey Juan Carlos, M\'ostoles, Madrid, Spain}

\address[label2]{Cytognos SL, Salamanca, Spain}

\begin{abstract}
Multicolor in situ hybridization (mFISH) is a karyotyping technique used to detect major chromosomal alterations using fluorescent probes and imaging techniques. Manual interpretation of mFISH images is a time consuming step that can be automated using machine learning; in previous works, pixel or patch wise classification was employed, overlooking spatial information which can help identify chromosomes. In this work, we propose a fully convolutional semantic segmentation network for the interpretation of mFISH images, which uses both spatial and spectral information to classify each pixel in an end-to-end fashion. The semantic segmentation network developed was tested on samples extracted from a public dataset using cross validation. Despite having no labeling information of the image it was tested on our algorithm yielded an average correct classification ratio (CCR) of 87.41\%. Previously, this level of accuracy was only achieved with state of the art algorithms when classifying pixels from the same image in which the classifier has been trained. These results provide evidence that fully convolutional semantic segmentation networks may be employed in the computer aided diagnosis of genetic diseases with improved performance over the current methods of image analysis.
\\
\\
\textit{This is the pre-peer reviewed version of the following article: "Pardo, E. , Morgado, J. M. and Malpica, N. (2018), Semantic segmentation of mFISH images using convolutional networks. Cytometry Part A", which has been published in final form at https://doi.org/10.1002/cyto.a.23375 . This article may be used for non-commercial purposes in accordance with Wiley Terms and Conditions for Self-Archiving.}
\end{abstract}


\end{frontmatter}


\section{Introduction}
Multicolor fluorescence in situ hybridization (mFISH) is a cytogenetic methodology that allows the simultaneous visualization of each chromosome pair in a different color, providing a genome-wide picture of cytogenetic abnormalities in a single experiment \citep{speicher1996karyotyping, schrock1996multicolor}. It was introduced in 1996 as spectral karyotyping (SKY) \citep{schrock1996multicolor} and multiplex-FISH (M-FISH) \citep{speicher1996karyotyping}, similar methodologies in terms of labeling but differing in terms of imaging system requirements and image acquisition and analysis process. After the mFISH spectral information has been acquired, different features can be analyzed to assign a chromosome label to each pixel. Manual interpretation of mFISH images is a time-consuming task where not only the intensity of each pixel is compared across channels but also the shape, size and centromere position. Many attempts were made to automate the task, being the most notable approaches pixel and region based classifiers. These classifiers usually build a feature vector using pixel or patch based intensity information and use that information to train a classifier, which is later used to classify pixels from the same image \citep{wang2017patch, li2012classification}, or from a different one \citep{choi2004joint, choi2008feature}.
Multiple pixel based classifiers have been developed for the analysis of mFISH images, showing that the spectral information present in a pixel can be successfully used to train machine learning classifiers. \citep{schwartzkopf2005maximum, choi2008feature}. In the other hand, region based classification has also been studied, showing that it generally outperforms pixel based classification approaches.\citep{li2012classification, wang2017patch}, underlining the importance of using spatial information to improve the performance of mFISH analysis algorithms.

Despite the relative success of the above mentioned approaches, none of them take into account spatial information about the shape, size, or texture of the objects being analyzed. This limits the performance of the algorithms in challenging scenarios where the identification of the chromosome is not clear based only on the spectral information. Some important features typically used in manual analysis, but not incorporated into classification algorithms, are the relative length of a chromosome, the arm ratio, or the centromeric index \citep{lejeune1960proposed}. Such features can be automatically learned running the input images through a network of convolutions and resampling operations, comparing the resulting image to the expected segmentation map, and backpropagating the error to learn the network parameter. This approach is usually called “end to end semantic segmentation”.

End to end semantic segmentation using convolutional networks has been shown to achieve state of the art results by automatically learning features based on spatial and intensity information \citep{ronneberger2015u, badrinarayanan2015segnet, chen2016deeplab}. The convolutional network approach shifts the focus from feature engineering to network architecture engineering, searching for the best network layout for a given problem. 

In the field of biomedical image processing, network architectures such as U-Net \citep{ronneberger2015u} have been widely used to perform end to end semantic segmentation. This architecture consists of two paths: the first one builds an abstract representation of the image by iteratively convolving and subsampling the image, while the second creates the target segmentation map by iterative upsampling and convolving the abstract feature maps. These two paths are symmetrical and connected by connecting each subsampling step with the analogous upsampling step by concatenating the corresponding layers. 

Different architectures of end-to-end convolutional networks for semantic segmentation have been developed since the creation of U-Net, being Deep-Lab architecture \citep{chen2016deeplab} \citep{chen2017rethinking} one of the best performing ones, with an average precision of 86.9\% in the Pascal VOC challenge \citep{everingham2010pascal}. The core of this architecture is the use of atrous convolution for probing convolutional features at different scales which has proven to be a powerful way of incorporating context information. The good results in the Pascal VOC 2012 semantic segmentation challenge led us to incorporate some of the main ideas into our work with mFISH images.

The main challenge of applying end to end convolutional networks is the limited number of samples found in the commonly used benchmarks, mainly the ADIR dataset \citep{choi2004joint, choi2008feature, li2012classification, wang2017patch}. This dataset contains mFISH samples prepared using Vysis, ASI, and PSI/Cytocell probes, were each cell is captured in 7 images, 6 of them representing the observed intensity of each fluorophore and the remaining one containing manual expert labeling of every pixel. The three different probes used to prepare the samples do not share a common labeling scheme, which means that the features used to segment a sample hybridized with a Vysis probe set may not work in samples where the ASI probes were used. To reduce the impact of using different probe sets for training and testing, this work focuses on the Vysis subset, since it is the largest one.

In this work, we present a fully convolutional network for semantic segmentation of mFISH images that uses both spectral and spatial information to classify every pixel in an image in and end-to-end fashion and provide evidence that our approach performs well even in challenging scenarios.

\section{Materials}

The ADIR dataset was used to design and evaluate the network. This dataset contains samples prepared with different probe sets: ASI, PSI/Cytocell and Vysis. Each probe set uses different dyes and combinatorial labeling schemes, this means that even if all mFISH images have 6 channels, these channels have different meanings depending on the probes used. There are some cases when the emitted spectra overlap among different subsets, ASI and Vysis probes both emit fluorescence in the Spectrum Green channel, and ASI and PSI/Cytocell both emit fluorescence in the Cy5 channel. Despite this overlap, the underlying probes are hybridized to different chromosomes, which means that this information is not easily reusable for learning the segmentation maps. The dataset contains 71 ASI images, 29  PSI/Cytocell images, and 84 Vysis images. We decided to evaluate our algorithm on samples prepared with the Vysis probe sets, since they are the most frequent.  

\begin{figure}[t!]
    \centering
    \begin{subfigure}[t]{0.5\textwidth}
        \centering
        \includegraphics[width=0.9\textwidth]{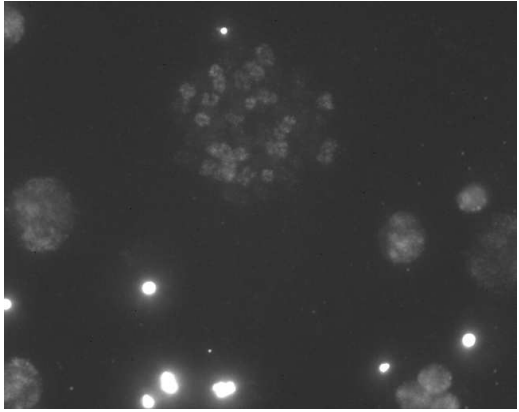}
        \caption{Far red channel of image V190442}
    \end{subfigure}%
    ~ 
    \begin{subfigure}[t]{0.5\textwidth}
        \centering
        \includegraphics[width=0.9\textwidth]{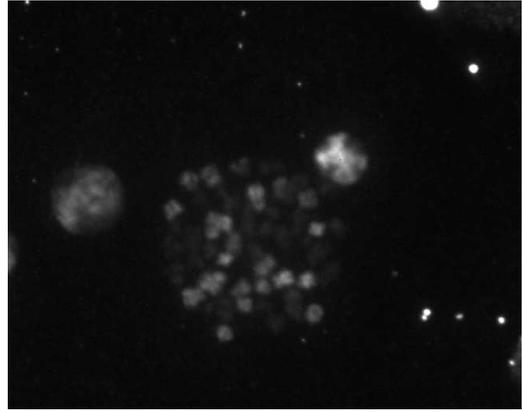}
        \caption{Far red channel of image V190542}
    \end{subfigure}
    \caption{The far red channel of the image V190442 has different intensity levels than other images from the dataset. The same channel extracted form image V190542 is shown for comparison.}
    \label{sample_issue}
\end{figure}

The Vysis subset was further refined by removing 14 low quality images. Some authors have reported a list of low quality images due to ill-hybridization, wrong exposure times, channel cross talk, channel misalignment, or using different probes that the ones reported \citep{choi2008feature}. To ensure that the estimated CCR is not biased by avoidable issues in the sample preparation and acquisition steps, the images listed in \citep{choi2008feature} have been removed from the dataset. Additionally, when analyzing the achieved CCR on the remaining samples we detected some outliers, visual inspection of these samples confirmed issues in the preparation or acquisition steps which can be observed in figure \ref{sample_issue}. We removed 4 additional samples that presented abnormal intensity levels in some of the channels and limited the performance of the network, we also kept samples with similar but less intense problems since most of the noisy samples did not have large negative impact to the performance of the network and helped to maintain a realistic variability in the dataset. The list of removed images can be consulted in table \ref{removed_images}.


\begin{table}[!htb]
\begin{center}
  \caption{List of images removed from the Vysis subset}\label{removed_images}
  \footnotesize
  \begin{tabular}{| l | l || l | l |}
    \hline
    File name & Condition & File name & Condition \\ \hline
    V250253 & Ill-hybridization/Wrong exposure & V290962 & Ill-hybridization/Wrong exposure \\ \hline
    V260754 & Channel cross talk & V291562 & Channel misalignment \\ \hline
    V260856 & Channel cross talk & V1701XY & Wrong probe label \\ \hline
    V290162 & Channel cross talk & V1702XY & Wrong probe label \\ \hline
    V290362 & Channel cross talk & V1703XY & Wrong probe label \\ \hline
    V270259 & Ill-hybridization/Wrong exposure & V1402XX & Ill-hybridization/Wrong exposure \\ \hline
    V280162 & Ill-hybridization/Wrong exposure & V190442 & Ill-hybridization/Wrong exposure \\ 
    \hline
  \end{tabular}
\end{center}
\end{table}

\section{Methods}

\begin{figure}[t!]
    \centering

    \includegraphics[width=1.0\textwidth]{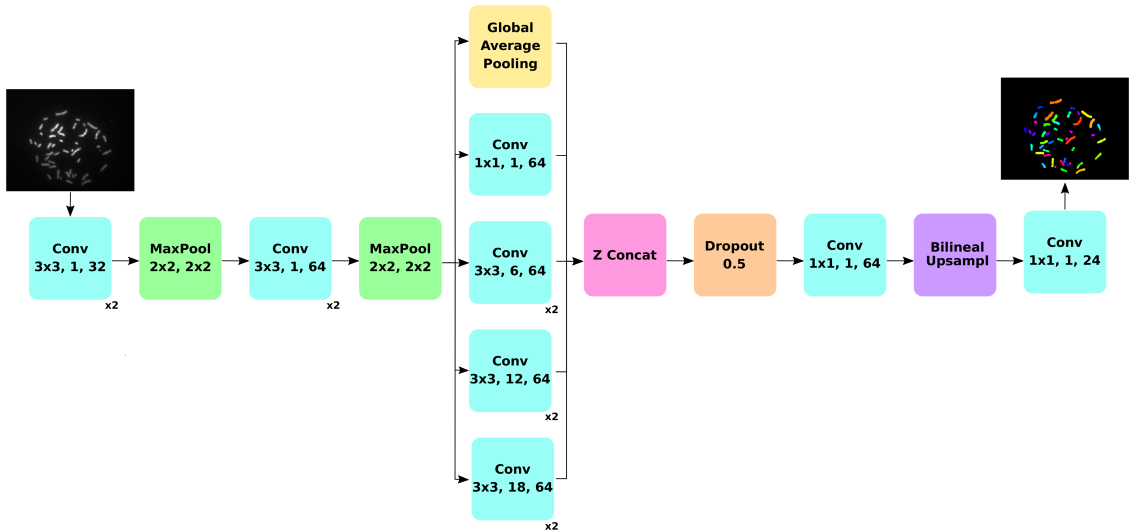}

    \caption{Network architecture. The Conv block illustrates a convolution followed by a ReLU activation and batch normalization, for the last Conv block there is no batch normalization and the activation is switched to a Softmax function. The parameters in a Conv block represent the kernel size, the dilation rate, and the number of filters. The MaxPool block represents a max pooling operation where the first parameter is the pool size, and the second is the stride. The parameter for the Dropout block represents the dropout rate. Whenever a $x2$ is present, 2 blocks are performed sequentially.
    After two downsampling steps, an ASPP module is used to probe information at different resolutions. The different ASPP branches are concatenated and a 1x1 convolution is used to combine the information. The final feature maps are upsampled using bilinear interpolation. We found useful to apply dropout after concatenating the ASPP branches due to the low number of samples available.}
    \label{network_architecture}
\end{figure}


The low number of samples in the ADIR dataset, their variability and the large number of classes to be segmented led to a set of carefully designed choices in the network architecture. The underlying driving force when designing the network was to use cost-effective convolutional blocks in terms of number of parameters and performance. Thus we have designed a network which is relatively shallow when compared to deeper ones such as ResNet \citep{he2016deep} or Inception \citep{szegedy2015going}, but nonetheless achieves high CCR in the segmentation of mFISH samples. This section presents the main blocks of the network, and describes the training procedure. An overview of the network architecture is illustrated in figure \ref{network_architecture}.



\subsection{Convolutional block}
Convolutional networks are usually comprised of different blocks that encapsulate a specific behavior. The VGG network \citep{simonyan2014very} uses a basic block in which 2 or more convolutions with 3x3 kernels are followed by a pooling operation, the idea behind this block is that a stack of two or more 3x3 convolutional layers, without spatial pooling between them, emulates a larger receptive field with a smaller number of parameters; as an example, two 1-channel 3x3 convolutional layers have 9 parameters and an effective receptive field of 5x5, whereas a 1-channel 5x5 convolutional layer has 25 parameters for the same receptive field.

Since the creation of VGG, deeper networks such as Inception \citep{szegedy2015going} and ResNet \citep{he2016deep} have been developed. These networks are usually trained on datasets containing thousands of images and are designed to account for the large variability present on those datasets. Specifically, the main idea behind Inception is to use dense components to approximate the local sparse structure usually found in convolutional networks. That is, to cover local activation clusters using 1x1 convolutions and more spatially spread activation clusters using larger 3x3 and 5x5 convolutions. On the other hand, ResNet blocks introduce shortcut connections to ease the training of deep networks. Because the deepest layers of a network introduce small changes to the features, residual learning is better preconditioned than standard learning. Despite this progress in convolutional networks, the relatively small size of the ADIR dataset makes these blocks unfit for the semantic segmentation of mFISH images.

Because the goal of this work is to build a cost effective convolutional network for the segmentation of mFISH images, a VGG-like layout was used in the first section of the architecture. The first 4 blocks in figure \ref{network_architecture} are comprised of 2 pairs of 3x3 convolutions followed by max pooling operations. This layout resembles the combination of small 3x3 kernels and downsampling operations used in the VGG network and its goal is to create an initial set of high level features that will be later refined by using dilated convolutions to aggregate contextual information.

\subsection{Dilated convolution}
Atrous or dilated convolution \citep{yu2015multi, chen2016deeplab, chen2017rethinking} is an efficient way of aggregating multiscale contextual information by explicitly adjusting the rate at which the input signal is sampled or, in an equivalent view, the rate at which the filters are upsampled. Specifically, this operator is a generalization of the standard convolution where the value of the dilated convolution between signal $f$ and filter $k$, with dilation rate $l$, is computed following equation \ref{dilated_convolution}.

\begin{align}
(f *_l k)(x) = \sum_{m = -\infty}^{\infty} f(m) k(x - l m)
\label{dilated_convolution}
\end{align}

By tuning the dilation rate, the network can probe image features at large receptive fields. This enables the network to include long range context information with a more limited cost than using successive convolution and pooling operations. On the other hand, it is easy to see that equation \ref{dilated_convolution} with a dilation rate $l$ of 1 is equivalent to a standard discrete convolution.

When comparing dilated convolution to the way U-Net style networks aggregate context information, one drawback of U-Net style architectures is that the context information is captured using downsampling operations. Pooling operations are useful to build high level features at the expense of losing resolution. This is very convenient for the classification task since no location information needs to be preserved. On the contrary, semantic segmentation performs a pixel wise classification, meaning that spatial information needs to be preserved however, the spatial information lost on the downsampling path is recovered by complex upsampling operations involving transpose convolutions, and regular convolutions. The upsampling path may be avoided if contextual information is captured using modules that do not downsample the feature maps. This is where dilated convolutions come into play.


The proposed architecture introduces dilated convolutions after the second downsampling operation. For the problem of mFISH semantic segmentation, we found that building a first level of abstract features using two downsampling operations, works better than using a deeper first stage with more downsampling operations or a shallower one with fewer ones.


\subsection{Atrous spatial pyramid pooling}
Spatial pyramid pooling \citep{he2014spatial} was designed to overcome the problem of analyzing information at different scales, sizes, and aspect ratios. The module pools image features at different scales to build a fixed size feature vector. This enables the classification of arbitrary sized images, and also improves the performance for objects undergoing deformations or scale transformations.

The successful application of spatial pyramid pooling in image classification and object detection led to the development of atrous spatial pyramid pooling \citep{chen2016deeplab,chen2017rethinking}. This new module applies some of the core ideas of spatial pyramid pooling to the field of image segmentation, dilated convolutions are used to capture information at different resolutions, 1x1 convolutions express the degenerate case of dilated convolutions where only the center weight is active, and global average pooling is used to capture image features.

In this work, the atrous spatial pyramid pooling module is introduced after the second max pooling operation. This module performs a resolution wise analysis of the initial set of low level features, aggregating spatial information to improve the initial spectral analysis.







\subsection{Training}
In order to train the network, a set of guidelines must be followed to fully reproduce our work. This section presents the key points of our training protocol.

\textbf{Loss function:} The training is performed in an end-to-end fashion, where a batch of 6 channel images is fed into the system and the network outputs a batch of 24 channel images representing the class likelihood for each pixel. The output of the network is converted into a categorical distribution using the softmax function \ref{softmax}, where $S_c(x)$ denotes the softmax value of class $c$ at pixel $x$, $f(x)_c$ is the value of the feature in the pixel $x$ and channel $c$, and $C$ represents the number of channels or classes. Finally, this categorical distribution is compared to the ground truth using the cross entropy loss function \ref{cross_entropy}, where $p(x)$ denotes the ground truth class distribution at pixel $x$ and $q(x)$ is the predicted distribution.

To compare the predictions to the ground truth, either the ground truth has to be scaled to the size of the logits \citep{chen2016deeplab}, or the predictions need to be scaled to the size of the ground truth \citep{chen2017rethinking}; scaling the ground truth would remove part of the information used in training, so the second option was chosen. Besides, the ground truth presents additional labels for background and overlapping chromosomes, however, these are not usually taken into account when training and calculating the CCR \citep{wang2017patch}, because of that, after creating the one hot encoded labels the image slices representing the background and overlapping labels are removed, this guarantees that the background and overlapping pixels are not taken into account during the training process.

\begin{align}
S_c(x) = \frac{e^{f(x)_c}}{\sum_{i = 1}^{C} e^{f(x)_i}}
\label{softmax}
\end{align}

\begin{align}
E(p, q) = -\sum_x p(x) log(q(x))
\label{cross_entropy}
\end{align}

\textbf{Optimizer:} The Adam algorithm \citep{kingma2014adam} was used to perform optimization. This optimizer was designed to combine the benefits of AdaGrad \citep{duchi2011adaptive}, which works well with sparse gradients usually found in computer vision problems, and RMSProp \citep{tieleman2012lecture} which works well in on-line and non-stationary settings. The method computes individual adaptive learning rates for different parameters from estimates of first and second moments of the gradients, and it has been shown to converge faster on some convolutional network architectures.

\textbf{Batch normalization:} Batch normalization \citep{ioffe2015batch} is used to regularize the features in the intermediate layers. When training convolutional networks, the inputs of a convolution layer have different distributions in each iteration. This is known as the internal covariate shift and is addressed by normalizing the inputs of a layer. This form of regularization improves the convergence and generalization of the network. Batch normalization works optimally when batches have a enough number of samples, so that the batch-wise statistics are significant. We decided to use a batch size of 16, since it has been shown to be sufficient for the segmentation network proposed in \citep{chen2017rethinking}. 

\textbf{Dropout:} Dropout \citep{srivastava2014dropout} was also used for regularization. This technique works by randomly setting to 0 a fraction of the units in a layer. While it has been widely used to create "thinned" fully connected layers when training classification networks \citep{krizhevsky2012imagenet, szegedy2015going}, it has also been used successfully in segmentation networks \citep{badrinarayanan2015segnet}. In this work, we introduced a  dropout layer between the concatenation of ASPP branches and the final 1x1 convolution, this forces the network to learn more significant and general features which, in turn, improves generalization. 

\textbf{Image preprocessing:} To speed up the process of training and enable larger batch sizes, the images were cropped and scaled. First, all samples were cropped to a 536x490 window, the minimum window that ensures that no chromosome information is left outside. The resulting images were then downscaled by 30\%, which produces images of 375x343 pixels.

\textbf{Data augmentation:} To prevent the network from over fitting the training samples we have used data augmentation. This is an essential step when training with a small number of samples since it increases the variance of the data used to train the network. Training samples were subjected to random scaling, rotations, and translations, which are some of the main types of deformations that can be observed in microscopic images, and the resulting images were added to the training set.










\section{Results}
The performance of the network is reported by estimating the CCR, computed using equation \ref{CCR}, of the Vysis samples from the ADIR dataset, using leave-one-out-cross-validation. The test was designed to avoid some common flaws encountered in the testing of mFISH classification algorithms, such as using a test set extracted from the same image the algorithm was trained on. In the following sections, we first analyze the performance of some state of the art methods and finally report the CCR achieved by our method.

\begin{align}
CCR = \frac{\#chromosome\ pixels\ correctly\ classified}{\#total\ chromosome\ pixels}
\label{CCR}
\end{align}

\subsection{Performance analysis of HOSVD}
We selected the HOSVD algorithm \citep{wang2017patch} to highlight the performance drop that some state of the art algorithms undergo when trained and tested on different images.

A branch of mFISH analysis algorithms including HOSVD \citep{wang2017patch, cao2012classification} are designed to perform analysis on the same image used for training. Specifically, HOSVD works by first selecting 30 random patches from every chromosome type in a image, these patches are later used to build the feature vectors needed to classify the rest of the patches in the same image. This analysis pipeline turns algorithms into semiautomatic approaches if the seed points are manually annotated, as in the case of the ADIR dataset. 

\begin{figure}[t!]
    \centering

    \includegraphics[width=0.99\textwidth]{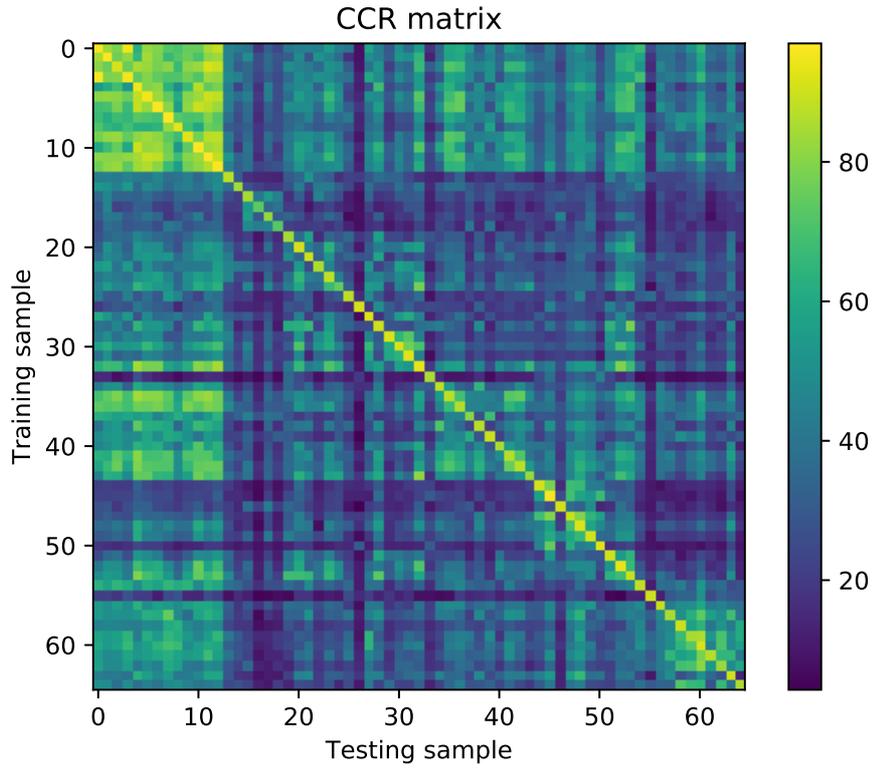}

    \caption{HOSVD error matrix.}
    \label{hosvd_error}
\end{figure}

The proposed dataset was used to evaluate HOSVD by testing every sample using each sample for training. Given a mFISH sample, 30 random patches for each chromosome type were used to build an HOSVD tensor, this tensor was later used to classify not only the rest of the image \citep{wang2017patch}, but also the rest of the dataset. This process was repeated for every sample in the dataset, and the CCR was computed at every iteration, generating the matrix illustrated in figure \ref{hosvd_error}. 

When performing training and testing on the same image, HOSVD achieved a CCR of 89.13\%, which is 2.49\% less than the CCR reported in the original work. In 98.46\% of the experiments, the highest CCR was achieved when performing training and testing on the same image. Only once did HOSVD achieve a higher CCR when training on a different sample to the one being tested. When excluding the sample being tested from the training set, the highest CCR averaged over all tested samples was 68.58\%, which is a 24.97\% less than the CCR reported in the original work. A similar performance reduction was reported in \citep{choi2008feature} for a different classification algorithm. In that case, the CCR dropped from 89.95\%, when performing self training-testing, to 72.72\%, when performing training and testing with different sets. These results show that state of the art performance is around 70\% for the analysis of unlabeled mFISH images.



\begin{figure}[t!]
    \centering
    \begin{subfigure}[t]{0.3\textwidth}
        \centering
        \includegraphics[width=0.9\textwidth]{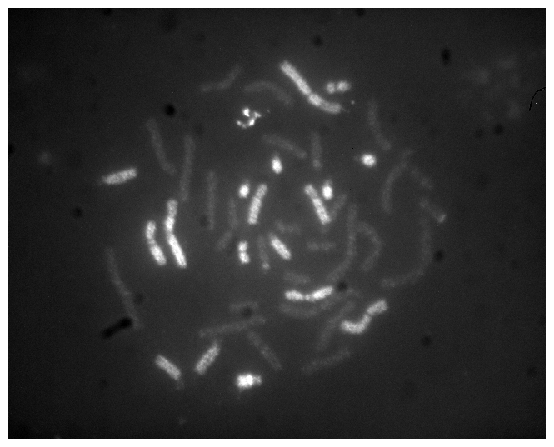}
        \caption{Aqua channel of image V1306XY}
    \end{subfigure}%
    ~ 
    \begin{subfigure}[t]{0.3\textwidth}
        \centering
        \includegraphics[width=0.9\textwidth]{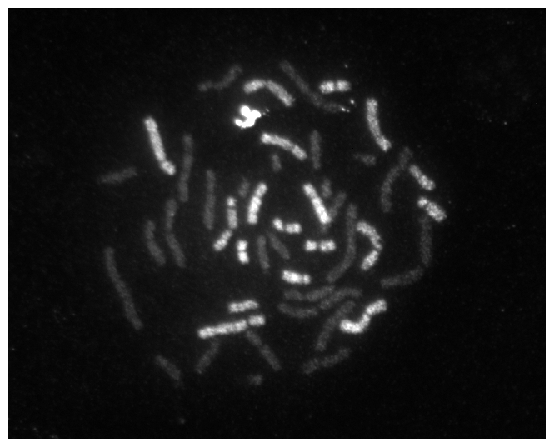}
        \caption{Far red channel of image V1306XY}
    \end{subfigure}%
    ~ 
    \begin{subfigure}[t]{0.3\textwidth}
        \centering
        \includegraphics[width=0.9\textwidth]{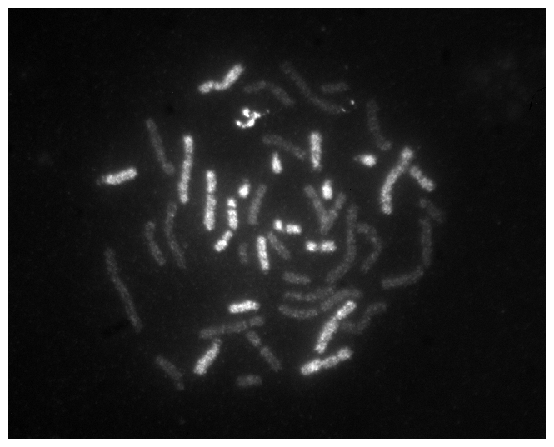}
        \caption{Green channel of image V1306XY}
    \end{subfigure}%
    ~ 
    
    \begin{subfigure}[t]{0.3\textwidth}
        \centering
        \includegraphics[width=0.9\textwidth]{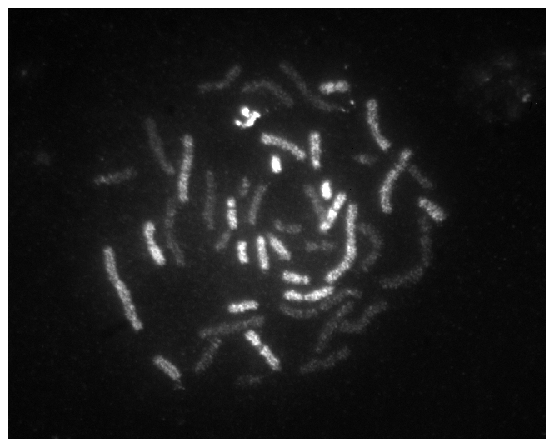}
        \caption{Red channel of image V1306XY}
    \end{subfigure}%
    ~ 
    \begin{subfigure}[t]{0.3\textwidth}
        \centering
        \includegraphics[width=0.9\textwidth]{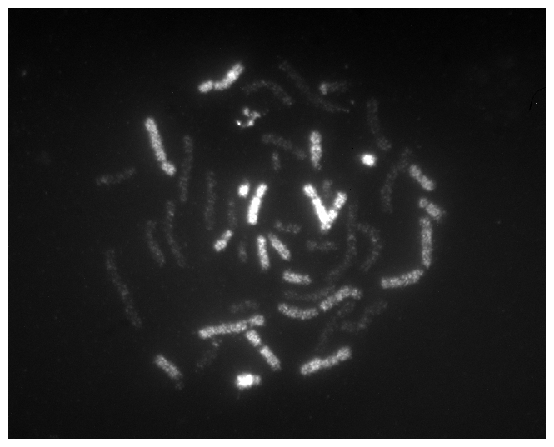}
        \caption{Gold channel of image V1306XY}
    \end{subfigure}%
    ~ 
    \begin{subfigure}[t]{0.3\textwidth}
        \centering
        \includegraphics[width=0.9\textwidth]{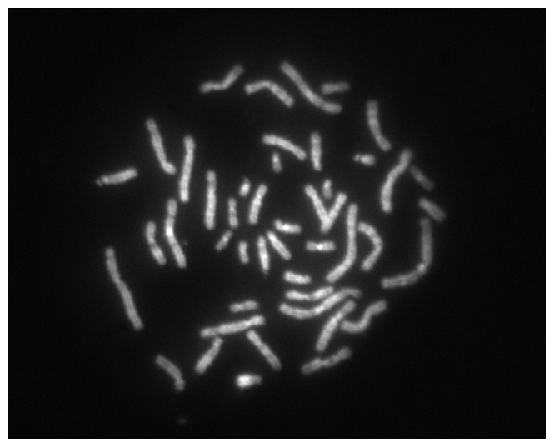}
        \caption{DAPI channel of image V1306XY}
    \end{subfigure}%
    ~ 
    \caption{Channels of V1306XY. All chromosomes present high intensity values in the DAPI channel, and some chromosomes are brighter than others in the rest of the channels.}
    \label{sample}
\end{figure}

\subsection{Results of semantic segmentation on the Vysis subset}

The 65 images in the dataset were used to train and evaluate the proposed architecture using leave one out cross validation. To estimate the CCR, the model was trained for 150 epochs and, for the last 5 iterations, the test set was evaluated, the final CRR estimate for the test set is calculated by averaging these CCR values. Following this methodology, the proposed method achieved a CCR of 87.41\%. To address the impact of the removed images the same evaluation procedure was carried for the whole Vysis subset, in this test the network achieved a CCR of 83.91\%.

\begin{figure}[t!]
    \centering
    \begin{subfigure}[t]{0.7\textwidth}
        \centering
        \includegraphics[width=1\textwidth]{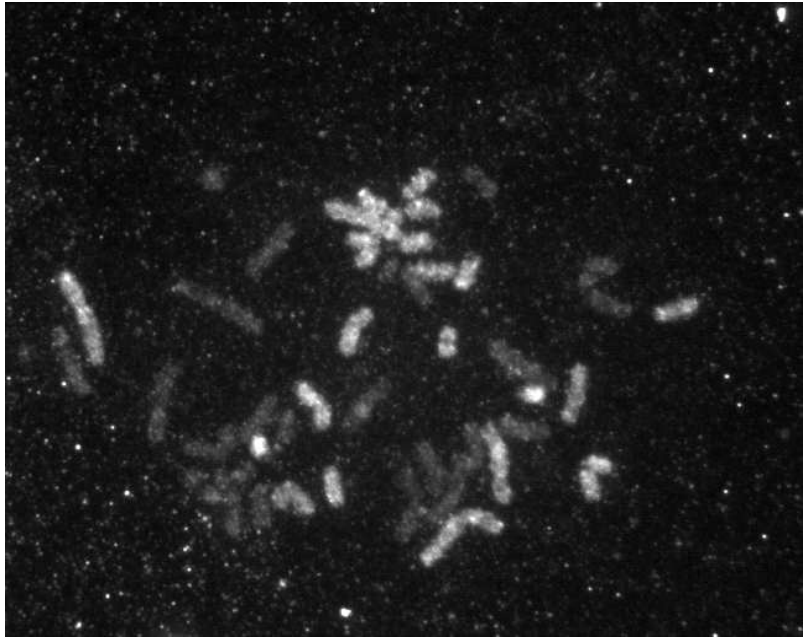}
        \caption{Far Red channel of V2704XY.}
    \end{subfigure}%
    
    \begin{subfigure}[t]{0.5\textwidth}
        \centering
        \includegraphics[width=1\textwidth]{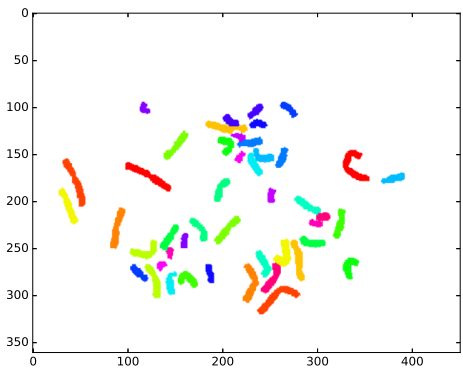}
        \caption{Ground truth}
    \end{subfigure}%
    ~ 
    \begin{subfigure}[t]{0.5\textwidth}
        \centering
        \includegraphics[width=1\textwidth]{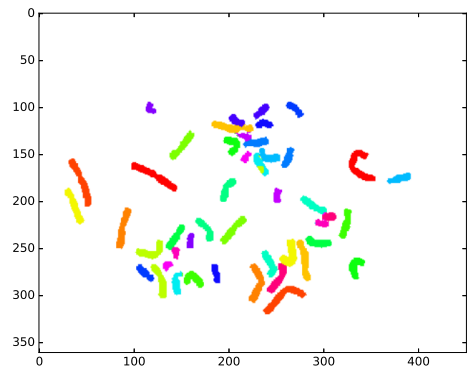}
        \caption{Prediction}
    \end{subfigure}
    \caption{The proposed network was applied to the sample V2704XY after being trained using the rest of the working dataset. The method achieved a CCR of 99\%.}
    \label{sample_results}
\end{figure}

\section{Discussion}

Our tests have shown that HOSVD underperforms, similarly to prior work, when analyzing unlabeled samples. Although the common approach in machine learning research is to build feature vectors using images other than the one being analyzed, this procedure seems to reduce the performance of state of the art algorithms when compared to the results achieved while training and testing on the same image.

Comparing the results achieved by HOSVD when performing classification and training on the same image to the results achieved by our approach suggests that, for the ADIR dataset, using HOSVD may be more robust to exposure variability across images. However, given that our method significantly outperforms state of the art algorithms on unlabeled samples, one can expect that end to end segmentation using convolutional networks will completely outperform algorithms that perform training and analysis on the same image.

For the sample shown in figure \ref{sample_results}, despite the presence of speckle noise, the proposed networks achieves a CCR of 99\% while HOSVD achieves a lower score of 90\%. This result may also suggest that, while our method is more susceptible to overall changes in the sample intensity levels than HOSVD, it is more robust to image noise and will achieve optimum results on larger and carefully acquired datasets.

The results also suggest that end to end convolutional networks exploit a richer set of features than previous algorithms. The analysis of both spectral and spatial features leads to a CCR increase of at least 20\% when compared to previous algorithms in the unlabeled image scenario, and a CCR drop smaller than 3\% when compared to HOSVD analysis with prior labeling information.

\section{Conclusion}
In this work, we proposed a convolutional network architecture for the semantic segmentation of mFISH images. The architecture shares some of the foundations of VGG \citep{simonyan2014very}, spatial piramid pooling networks \citep{he2014spatial}, dilated convolution networks \citep{yu2015multi}, and the DeepLab architecture \citep{chen2017rethinking} while adapting them to the field of mFISH semantic segmentation. VGG blocks build an initial set of low level features, and dilated convolutions further refine them following a multi resolution strategy, a pyramid pooling layout is used used to capture context at several ranges and the information is combined using a concatenation + dropout strategy. The final feature set is upsampled using bilineal interpolation resulting in the final segmentation map.


Our experimental results show that the proposed algorithm achieves state of the art CCR for the analysis of unlabeled images. Our end to end architecture scored a CCR of 87.41\% in the Vysis subset of the ADIR dataset, which is a 27\% better than HOSVD results when classifying images that were not used in training time, and a 20\% better than the results reported in \citep{choi2008feature} when using a subset of the testing image set for training. These results underline the importance of using end to end architectures to further exploit spatial information while leveraging the rich spectral information available when training on multiple images.

Finally, the number of samples and, specially, the relation between number of samples and number of classes may be a limiting factor of this approach. Successful applications of end to end convolutional networks are usually trained on thousands of samples. For this reason, we believe that training with a larger sample size will improve the CCR and allow for deeper networks that have been successfully used in the semantic segmentation of other image sets.



\section{Acknowledgments}
This work was partially funded by Banco Santander and Universidad Rey Juan Carlos in the Funding Program for Excellence Research Groups, ref. "Computer Vision and Image Processing" and by Project RTC-2015-4167-1 of the Spanish Ministry of Economy and Competitiveness. We gratefully acknowledge the support of NVIDIA Corporation with the donation of the Tesla K40 GPU used for this research.

\bibliographystyle{model1-num-names}\biboptions{authoryear}

\bibliography{sample.bib}

\end{document}